\documentclass[12pt]{article}

\usepackage{sbc-template}
\usepackage{graphicx,url}
\usepackage[utf8]{inputenc}
\usepackage[english]{babel}
\usepackage{subcaption} 
\usepackage{algorithm} 
\usepackage{algpseudocode} 
\usepackage{hyperref}
     
\title{LLM-Driven Intrinsic Motivation for Sparse Reward Reinforcement Learning}
\author{André Quadros\inst{1}, Cassio Silva\inst{1} and Ronnie Alves\inst{1,2}}

\address{ Universidade Federal Pará, Brazil
  (UFPA)\\
  Belém -- PA -- Brasil
\nextinstitute
  Instituto Tecnológico Vale, Brazil (ITV) \\
  Belém -- PA -- Brasil
  \email{\{andre.rosario31, cassio266\}@gmail.com, ronnie.alves@itv.org}
}

\begin{document} 

\maketitle

\begin{abstract}

This paper explores the combination of two intrinsic motivation strategies to improve the efficiency of reinforcement learning (RL) agents in environments with extreme sparse rewards, where traditional learning struggles due to infrequent positive feedback. We propose integrating Variational State as Intrinsic Reward (VSIMR), which uses Variational AutoEncoders (VAEs) to reward state novelty, with an intrinsic reward approach derived from Large Language Models (LLMs). The LLMs leverage their pre-trained knowledge to generate reward signals based on environment and goal descriptions, guiding the agent.
We implemented this combined approach with an Actor-Critic (A2C) agent in the MiniGrid DoorKey environment, a benchmark for sparse rewards. Our empirical results show that this combined strategy significantly increases agent performance and sampling efficiency compared to using each strategy individually or a standard A2C agent, which failed to learn. Analysis of learning curves indicates that the combination effectively complements different aspects of the environment and task: VSIMR drives exploration of new states, while the LLM-derived rewards facilitate progressive exploitation towards goals.
\end{abstract}

\section{Introduction}

 Imagine that you are in a physical labyrinth, but through the intellectual challenge of reinforcement learning in environments where guidance is scarce. Imagine an agent, dropped into a complex maze, with only one piece of feedback: 'You've reached the end!' – and that feedback might only come after thousands of wrong turns. This, in essence, is the challenge of sparse rewards, a pervasive bottleneck preventing intelligent agents from learning efficiently in real-world scenarios. Our research confronts this fundamental dilemma: how do we empower an agent to effectively explore a vast, unfamiliar space to find those elusive rewards, and then reliably exploit that knowledge to consistently achieve its goals? This is the heart of the exploration-exploitation trade-off, a cornerstone of reinforcement learning. Traditional methods often struggle here. Random exploration is inefficient, and purely greedy exploitation can lead to getting trapped in suboptimal local optima. We hypothesized that for an agent to truly thrive in these challenging conditions, it needs more than just trial and error; it needs a richer understanding of its environment and a more informed way to generate and evaluate its actions.

Trying to learn through trial and error, without any confirmation that our action was positive or negative, can make any learning nearly impossible. This brings us to the problem of sparse environments. Knowing that there are environments where rewards can be distributed infrequently or extremely sparse makes these types of environments challenging for the agent \cite{devidze2022exploration}. Sparse rewards can be defined as a series of rewards produced through the interaction of the virtual reality agent with the environment in which the majority of the rewards received are not positive \cite{barto2012intrinsic}. The absence or lack of rewards that would have the purpose of guiding the actions performed by the agent causes a considerable increase in the difficulty of learning for the agent. As a consequence of the problem of sparse rewards, we have the problem of sampling efficiency, where, due to the scarce signals to guide learning, the agent ends up needing many interactions to learn. The exploration-exploitation dilemma is a fundamental challenge in Reinforcement Learning (RL). The Exploration-Exploitation Dilemma in RL, is the tension between trying new actions and search new states (exploration) to discover potentially better rewards, and sticking with actions that are already known to yield good results (exploitation) \cite{sutton2018reinforcement}. On the other hand, we have intrinsic motivation, a term from psychology, which in machine learning is described as learning carried out by an agent using an internal stimulus to learn new skills or achieve a goal \cite{barto2012intrinsic}. With intrinsic motivation, the agent generates intrinsic rewards which are independent of the environmental reward and which encourage exploration and taking actions in the environment.

The field of Natural Language Processing (NLP) has undergone a revolution since the introduction of the Transformer architecture \cite{vaswani2017attention}. This advancement paved the way for the development of Large Language Models (LLMs), which already have reference points with the development of large known and used models, such as Llama \cite{grattafiori2024llama} and Gemini \cite{team2024gemini}. Due to the colossal amount of information with which these models are trained, these models end up being imbued with a certain amount of world knowledge \cite{yu2023kola}. This enables them to demonstrate their capacity for the analysis and understanding of information \cite{cao2024survey}. The use of LLMs in RL is a field that explores this power of analysis and compression of LLMs to help with problems encountered in the field of RL.

To improve sampling efficiency in environments with sparse rewards, we created a combination of two methods for generating intrinsic motivation, applying the variational state as intrinsic motivation, using the KL-Divergence present in the loss calculation as an intrinsic reward, for new states. Adding an LLM as another intrinsic reward to guide the agent based on the importance of what is present in the state in relation to the task to be accomplished. Leveraging VSIMR's capability to explore new states and the LLM's knowledge to directly guide toward the environment's goal. One for exploring new states and another for finding, in states, those who will help to achieve the goal.

In \hyperref[sec:background]{Background section}, we will provide general information about intrinsic motivation and its utility in sparse reward problems. In \hyperref[sec:related]{Related Work}, we will present two experimental studies that together provided the basis for the RL strategies introduced in this work. We examine \hyperref[sec:methodology]{methodology} of the joint application of these techniques, along with the step-by-step process followed for the experiments. We analyze all results in \hyperref[sec:results]{Results and Discussion section} and finally in \hyperref[sec:conclusion]{Conclusions section}, we present our conclusions and future work suggestions.

\section{Background}
\label{sec:background}

\subsection{Reinforcement Learning}
Reinforcement Learning (RL) is a Machine Learning (ML) paradigm that seeks to learn through interaction to achieve a goal \cite{sutton2018reinforcement}. This learning occurs through trial and error, the learner who makes the decisions is called an agent. What the agent interacts with, understanding everything outside the agent, is the environment. They interact continuously, the agent selecting actions and the environment responding to these actions and presenting new situations to the agent. The environment returns signals and rewards with each interaction. The agent maximizes these rewards to learn \cite{sutton2018reinforcement}. The reward system is very important for the agent. Depending on the reward system, coming from the environment, the agent can learn to achieve its goal faster or not. The reward that the environment returns to the agent when it takes an action in it or coming from objects, external to the agent, is called extrinsic reward \cite{aubret2019survey} \cite{barto2012intrinsic}. The stimuli generated by extrinsic rewards coming from objects or the environment external to the agent are what will guide him to complete the task.

\subsection{Intrinsic Motivation}

Originally rooted in psychology, intrinsic motivation (IM) describes an individual's inherent drive to seek novelty and challenges, independent of external pressures \cite{ryan2000self}. This concept, informed by studies on early childhood learning \cite{piaget1952origins}, highlights a natural inclination in organisms to explore their surroundings and acquire new skills. Essentially, IM examines how an individual or agent's innate curiosity propels them to investigate an environment or problem, leading to new experiences and skill development without an explicit, pre-defined goal; rather, the goal is often achieved as a byproduct of this exploration.

In the realm of Reinforcement Learning (RL), various strategies attempt to replicate IM principles in learning agents \cite{aubret2019survey}. IM in RL can generally be categorized into two types: Knowledge acquisition and Skill learning \cite{aubret2019survey}. Knowledge acquisition  approach focuses on enabling the RL agent to learn about its environment. Models in this category enhance an agent's exploration efficiency and improve state abstraction. The Skill learning approach agents with the ability to abstract and master skills, either to achieve a specific goal or to build a reusable repertoire of capabilities for overcoming environmental challenges.
For this work, we have opted to concentrate on knowledge acquisition. Due to its proven effectiveness in encouraging the exploration of infrequently visited states within RL environments \cite{aubret2019survey}.
%
%
In particular, intrinsic motivation models categorized under knowledge acquisition have demonstrated the ability to significantly improve agent performance in sparse reward environments \cite{zahavy2020self}. The subsequent section will show the works used in the formulation of this approach.

\section{Related Work}
\label{sec:related}

\subsection{Variational State as Intrinsic Reward (VSIMR)}

The paper Variational State as Intrinsic Reward \cite{klissarov2019variational}, proposes a formulation of intrinsic motivation based on the Bayesian definition of surprise, in which experiences that deviate from the model's default choices are seen as surprising and specifically useful for learning. The idea is that the agent identifies states that cause important changes in its prior knowledge by measuring the difference between the posterior and prior distributions after visiting them. This measure of surprise can be used as an intrinsic application.
To implement this measure of surprise, a Variational Auto-Encoder (VAE) is used. The VAE projects the state space ($S$) into a latent probabilistic representation ($Z$) that represents the specific structure of the environment \cite{klissarov2019variational}. Importantly, the VAE maintains an approximate posterior distribution, $q\theta(Z|S)$, over this latent structure and uses a prior $p(Z)$ (typically a unitary Gaussian). The measure of the agent's surprise is obtained "naturally" from the VAE by quantifying how much the approximate posterior distribution $q\theta(Z|S)$ deviates from the prior $p(Z)$. This difference is measured by the KL-divergence: $r_{intrinsic-vae}(S) = KL(q\theta(Z|S)||p(Z))$. Providing this KL-divergence as an intrinsic reward encourages the agent to visit surprising regions of the state space.

\subsection{LLM to Reward Design}

In his paper LLM-Enhanced Reinforcement Learning: A Survey. Cao et al. present a catalog and classification of several ways in which LLMs are being used to support the RL field in a variety of ways. LLMs, with their vast knowledge and reasoning capabilities, emerge as promising reward designers in LLM-enhanced RL \cite{chakraborty2023re}, helping to create more effective reward functions \cite{ma2023eureka}. An approach to using LLMs as implicit reward designers (directly providing reward values) is direct prompting. This involves presenting LLM with natural language representations, such as the current state of the environment or examples of desired behaviors, and asking it to provide a reward signal based on its understanding. This technique can be used to extract auxiliary personality goals from general language, generating an intrinsic reward signal.

\section{Methodology}
\label{sec:methodology}

\begin{figure}[ht]
    \centering
    \includegraphics[width=0.3\textwidth]{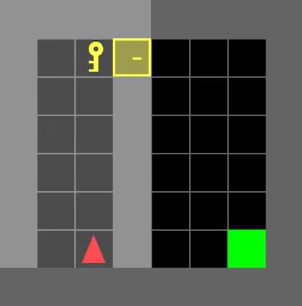}
    \caption{Door Key Environment}
    \label{fig:exampleFig1}
\end{figure}

To perform the proposed experiments in a sparse reward environment, we must choose an environment that has a good degree of difficulty, low frequency of rewards, and where the goal and its states can be easily described in words, since we will use an LLM to analyze the states of the environment. Minigrid \cite{MinigridMiniworld23} contains a collection of discrete grid-world environments to conduct research on Reinforcement Learning. Among these environments we have Door Key(Figure \ref{fig:exampleFig1}), which in its MiniGrid-DoorKey-8x8-v0 configuration, has a size of 8 squares high and 8 squares wide. This environment has a key that the agent must pick up in order to unlock a door and then get to the green goal square. The difficulty in this environment increases due to the fact that the environment only gives a reward to the agent when it manages to reach its final goal. In other words, the agent does not have any stimuli from the environment during its actions in the environment, not even in the actions that are crucial to achieving the goal, such as picking up the key and opening the door. Another difficulty in this environment is the agent's visualization. The agent has a partial observation of the environment, only seeing up to three squares to the sides and 7 squares in front, not seeing beyond walls and closed doors. The reward given to the agent when he reaches the objective also reflects his performance and is another difficulty. The extrinsic reward is inversely proportional to the number of steps the agent needed to achieve the objective. The fewer steps the agent uses to reach the objective, the greater is his reward. Due to the difficulty of this environment of sparse rewards and the ease of describing the objects present in the environment, this environment is perfect for use in our tests.

In this paper we use the Advantage actor critic (A2C) algorithm as the base RL algorithm. The A2C uses a hybrid design that combines the best features of policy-based (Actor) and value-based (Critic) methods using the moments when the algorithm performs advantage calculations to train the VAE.

Since the environment generates an extrinsic reward only when the agent reaches the goal, we will use the agent's intrinsic reward to guide it towards the goal. Using two combined intrinsic motivation strategies VSIMR and intrinsic reward based in LLM.

VSIMR, which gathers a set of states to train a VAE, and thus obtain a latent representation of the states, then calculates the distance between the probability distribution of a state S in relation to its latent representation Z, the KL Divergence, this distance is used as an intrinsic reward generated by the VAE. The KL Divergence value needs to be normalized before we use it as a reward. Thus we obtain the $r_{intrinsic-vae}$ of the current state.

\begin{figure}[ht]
\centering
\includegraphics[width=.7\textwidth]{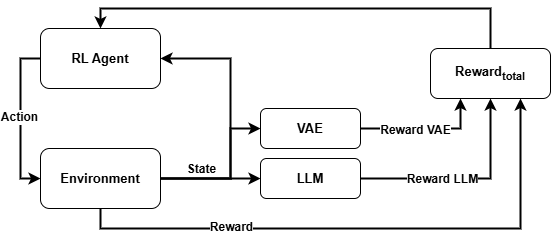}
\caption{Representation of the LLM+VAE strategy}
\label{fig:representation}
\end{figure}

In the intrinsic reward based on LLM, we use the Reward Design approach called Direct Prompting described in Cao et al.'s survey \cite{cao2024survey}, this approach is based on direct prompting with language descriptions. To simplify the use of this approach, we chose a popular model already trained, LLaMA 3.2 \cite{grattafiori2024llama}. With the chosen LLM, we need to think about how we will approach the creation of the prompt that we are going to generate and think about the response we want to obtain. Since we want an intrinsic reward and that, by concept, intrinsic motivation is an internal force that drives an individual to perform an activity for pure pleasure, interest, personal satisfaction, or challenge \cite{ryan2000self}. And the role of the VSIMR framework is curiosity and exploration of new states. We direct the prompt to evaluate how much a new state will help the agent achieve its goal. At each interaction of the agent with the environment, the agent will generate an intrinsic reward with the LLM. After the agent executes the action, the algorithm will format a prompt based on the state $s_{t+1}$. The prompt begins by explaining that the agent is observing an environment where it is located and reports that it has an objective which is the description of the environment's mission, and this is given by the environment by default, and then says which objects it is seeing in the new state, and finally asks on a scale of 0 to 10 whether or not the new state helps the agent in its final objective given in the mission. The final prompt is:

At the end the final reward that A2C will receive is the reward $r_{total}$, to guide the agent in the environment. The total reward is composed of the sum of the intrinsic reward, given by the environment, of the $r_{intrinsic-vae}$ which is the intrinsic reward generated by the VSIMR framework at each step with its regulator $\beta_{vae}$ and of the $r_{intrinsic-llm}$ intrinsic reward generated from the prompt at each step in the environment, also regulated by a hyperparameter $\beta_{llm}$.

\begin{equation}
     r_{total} = r_{extrinsic} + \beta_{vae} . r_{intrinsic-vae} + \beta_{llm} . r_{intrinsic-llm}
     \label{eq:rtotal}
\end{equation}

 We visualize in Figure \ref{fig:representation} the representation of the proposed strategy, demonstrating the RL Agent that takes an action in the environment that in turn generates a new state, which is processed by the agent through the VAE and LLM, the three generated rewards are applied in the Equation \ref{eq:rtotal}, to generate the total reward, which is the signal that will guide the agent. In  Algorithm 1, we have the A2C training loop implemented. For each training episode, we initialize a dataset $D$ where the state action tuples, the state $s_{t+1}$, the extrinsic reward $r_{extrinsic}$ and the intrinsic reward of the VAE $r_{intrinsic-vae}$ will be stored. We also initialize a dataset that will store the pairs of prompts made for a state $s_{t+1}$ and the value of the response given by the LLM. The dataset stored in $D$ will be used to train the VAE every N steps, and the data stored in $PromptReward$ will be used to avoid making repeated prompts to the LLM, and thus a large amount of processing time. Since we first consult this dataset and if the prompt has never been processed, then we send it to the LLM and store this new one for future queries. At each step performed by the agent, it takes an action based on the A2C algorithm policy and receives a new state $s_{t+1}$ and an extrinsic reward $r_{extrinsic}$ given by the environment. We compute the KL Divergence with the VAE, which returns a normalized value that is our intrinsic reward from the VAE $r_{intrinsic-vae}$, then we store in $D$ the tuple with the data needed to train the VAE.
To generate the intrinsic reward from an LLM, we follow similar steps: we generate the prompt from the new state $s_{t+1}$, and with the prompt ready, we check in the dataset $PromptReward$ if the question has already been asked and if we already have a reward for that question. If it has not yet been asked and is not in the dataset, we process the prompt in the LLM, store the prompt and answer pair in the dataset, and we have our intrinsic reward generated from the language model, the $r_{intrinsic-llm}$. We use the timestep that the A2C trains the actor and the critic, to also train the VAE in the last set of states. And so we continue with our loop until the algorithm ends.

\begin{algorithm}
    \caption{Intrinsically motivated training loop for A2C using VAE and LLM}
    \label{alg:encontrar_maior}
    \begin{algorithmic}[1]
    \For{Episode=0,1,2, n}
        \State Initialize the dataset $D$ and insert $s_{0}$ in $D$;
        \State Initialize the dataset $PromptReward$;
        \For{t=0,1,2...T}
            \State Take an action and watch the next state $s_{t+1}$ and the extrinsic reward $r_{extrinsic(st+1)}$\;;

            \State Compute $r_{intrinsic-vae(st+1)}$ (st+1) = KL Divergence\;;            
            
            \State Store tuple ($s_{t+1}$, $a_{t}$,$r_{extrinsic(st+1)}$, $r_{intrinsic-vae(st+1)}$) in $D$;

            \State Compute $r_{intrinsic-llm(st+1)}$ (prompt(st+1)) = LLM Answer;
            \State Store tuple (prompt(st+1), LLM Answer) in $PromptReward$;

            \If{mod(t,N) = 0}
                \State Train Actor and Critic  with the return Gt=$\Sigma_{t}r_{extrinsic(st)}+$ $ \beta_{vae} r_{intrinsic-vae(st+1)}$ + $ \beta_{llm} r_{intrinsic-llm(st+1)}$\;;

                \State Train VAE with the states $s$ collected in $D$\;;

                \State Initialize the dataset $D$ and insert $s_{t}$ in $D$\;;

            \EndIf
        \EndFor
    \EndFor
    \end{algorithmic}
\end{algorithm}

In the next section we will see the results of this algorithm and discuss about its benefits and drawbacks.

\section{Results and Discussion}
\label{sec:results}

\begin{figure}[ht]
    \centering 

    \begin{subfigure}[b]{0.7\textwidth} 
        \centering
        \includegraphics[width=\textwidth]{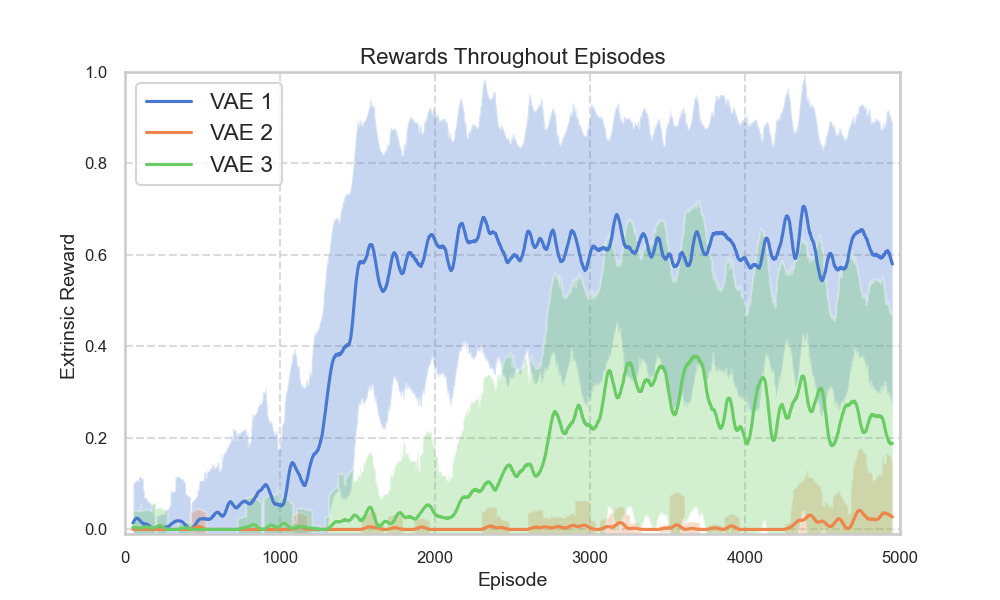}
        \caption{Results do VAE}
        \label{fig:resultvae}
    \end{subfigure}
    \hfill 
    \begin{subfigure}[b]{0.7\textwidth} 
        \centering
        \includegraphics[width=\textwidth]{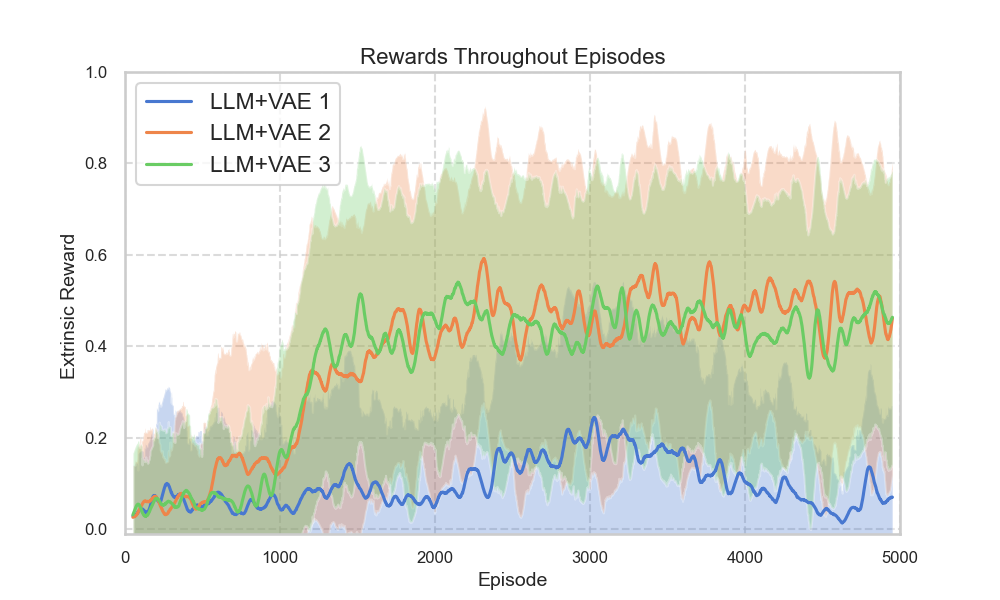}
        \caption{Results LLM+VAE}
        \label{fig:resultllmvae}
    \end{subfigure}

    \caption{Comparison of results in Door Key Environment: (a) using VAE, (b) using LLM+VAE.} 
    \label{fig:door_key_comparison} 
\end{figure}

To apply the methodology, we ran Algorithm 1 in the DoorKey environment. The MiniGrid-DoorKey-8x8-v0 configuration used gives us 640 steps by default to reach the goal. However, to better evaluate the results, we increased this value by 40 percent, reaching a total of 896 steps, all other environment settings remain at default. We prepared a method in the code to translate the descriptive values of the state $s_{t+1}$ into a prompt following the query pattern mentioned earlier in this paper. The code applying the algorithm and which generated such results can be found on \href{https://github.com/andre3103/LLM-Driven-Intrinsic-Motivation-for-Sparse-Reward-Reinforcement-Learning}{github}\footnote{The repository with the implementation can be found at: https://github.com/andre3103/LLM-Driven-Intrinsic-Motivation-for-Sparse-Reward-Reinforcement-Learning}. To properly evaluate the hypothesis raised, that the use of LLM+VAE in the intrinsic motivation of an agent can help it fulfill its mission in sparse environments, we ran the algorithm 3 times with VAE only and the algorithm using LLM+VAE. Sparse reward environments can generate different results from the same algorithm in the same environment, each time it is executed. That's why we run each algorithm 3 times and smooth the results with the mean and standard deviation of each execution, so we can perform a better analysis of these executions.These results are shown in Figure \ref{fig:resultvae} and  Figure \ref{fig:resultllmvae}.

Looking at the results in Figure \ref{fig:resultvae}, the vae 1 agent shows the most successful learning curve among all the experiments. It quickly learns to achieve high rewards, consistently reaching average rewards above 0.6, often close to 0.7 and even peaking near 0.8. The variance is also high, indicating strong performance but with some fluctuations. This suggests excellent exploration early on, followed by effective exploitation. Demonstrating the VAE's great ability to explore the environment efficiently \cite{klissarov2019variational}. The second VAE agent performs very poorly, achieving consistently low rewards near zero throughout the entire training period. It seems completely unable to learn. At last the VAE 3 shows a moderate learning curve, eventually reaching average rewards between 0.2 and 0.4, with some peaks. It performs better than VAE 2 but significantly worse than VAE 1.

When analyzing the performance of agents using LLM+VAE (Figure \ref{fig:resultllmvae}) ; The LLM+VAE 1 shows very low total rewards, consistently staying below 0.2, and often near zero, even after 5000 episodes. The variance is also relatively low, suggesting it consistently performs poorly. This agent appears to be largely failing to learn or is stuck in a local optimum that yields minimal rewards. It likely struggles with both exploration (not finding paths to rewards) and exploitation (not consistently leveraging any rewards it might stumble upon). The second LLM+VAE agent shows a significant improvement compared to LLM+VAE 1. It starts learning around episode 1000 and reaches average rewards between 0.3 and 0.5, with peaks around 0.6. There's considerable variance, suggesting some runs perform better than others, but overall it demonstrates a capacity to find and exploit rewards. The LLM+VAE 3 performs similarly to LLM+VAE 2 in terms of average reward, also reaching between 0.3 and 0.5, and even slightly higher peaks than LLM+VAE 2 at times. Its learning curve seems to follow a similar trajectory and like the LLM+VAE 2, it shows good exploration and exploitation.

\begin{figure}[ht]
\centering
\includegraphics[width=.9\textwidth]{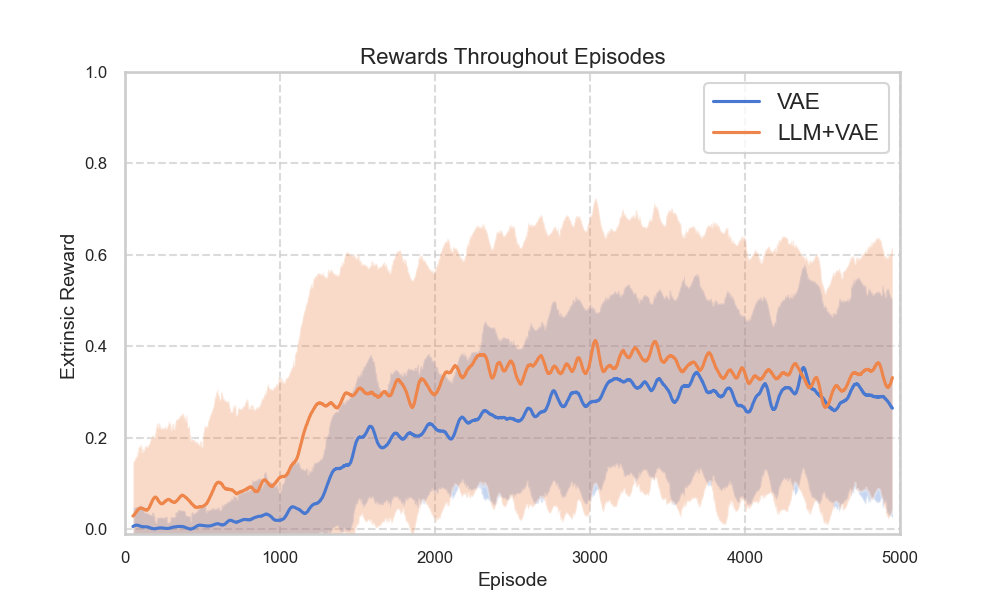}
\caption{Overlap of means and standard deviations of results.}
\label{fig:overlapResults}
\end{figure}

In Figure \ref{fig:overlapResults}, we show the mean and standard deviation of the three executions of each algorithm. Knowing that the environment with such sparse rewards can promote such different results, we combined the three executions of each algorithm, VAE and LLM+VAE, in order to reach a definitive conclusion about the performance and results of the algorithms. Considering that in the executions of Figures \ref{fig:resultvae} and \ref{fig:resultllmvae} there are very different variations and performances between the executions of the same algorithm and the competing algorithm.
VAE algorithm shows a modest learning curve, reaching average rewards around 0.2-0.3, with some peaks up to 0.4. Otherwise the LLM+VAE consistently outperforms the VAE baseline in this specific aggregated comparison, achieving average rewards between 0.3 and 0.4, with peaks closer to 0.5. The learning curve is also slightly steeper initially. Demonstrating that the combination of techniques shows a good improvement, both in the speed of learning, observed that LLM+VAE shows superior average rewards from the beginning, and in the stability of learning, as seen in the greater stability of LLM+VAE compared to VAE, demonstrating that the algorithm managed to reach the objective of the environment with fewer steps than VAE.


Our experiments revealed a nuanced picture of agent performance. Some configurations, such as LLM+VAE 1 and VAE 2, frequently converged to suboptimal local optima, exhibiting significant challenges in effective exploration and consequently failing to discover paths to rewards within the sparse environments. This suggests that while these components hold promise, their integration or specific hyperparameter settings might impede thorough environmental exploration.

Conversely, agents like VAE 1, and crucially, LLM+VAE 2 and 3, demonstrated successful exploration, reliably identifying rewarding trajectories, followed by efficient exploitation to consistently achieve higher rewards. The LLM+VAE technique, in particular, showcased this superior performance in two out of three executions, highlighting its potential for more robust exploration and exploitation when properly configured. However, a direct comparison between individual runs of LLM+VAE and VAE alone revealed considerable variability. While LLM+VAE 2 and 3 achieved strong results, LLM+VAE 1 represented a near-complete failure. Similarly, VAE 1 emerged as the top performer across all experiments, yet VAE 3 exhibited the worst performance. This marked instability in VAE-only configurations underscores the challenges in relying solely on VAEs for consistent performance in sparse reward settings.

\section{Conclusions}
\label{sec:conclusion}

Looking ahead, several avenues for future research emerge from this study. First, the inconsistent performance of some agents, particularly the failures in exploration, strongly indicates a need for hyperparameter fine-tuning. Future work will involve a more systematic and extensive exploration of the hyperparameter space for both LLM+VAE and VAE configurations to optimize their exploratory capabilities. Second, given the LLM's role in guiding exploration, a critical area for investigation is the construction of prompts and queries fed to the LLM. We hypothesize that crafting more effective prompts will enable the LLM to provide superior guidance, leading to better answers and, consequently, more efficient exploration by the RL agent. A deeper analysis into the impact of prompt engineering on LLM-driven exploration is warranted. Furthermore, once rewarding states are identified, a more profound understanding of the environment, facilitated by the LLM, could pave the way for more efficient exploitation. Future research will delve into how the LLM can further enhance the agent's ability to consistently achieve higher rewards once rewarding paths are discovered. This could involve using the LLM for refined policy adjustments or improved value function estimatation.

The core hypothesis driving our LLM integration was to enhance exploration by leveraging the LLM's potential for better state representations or more intelligent action proposals. The higher average rewards observed in the successful LLM+VAE runs lend credence to this idea, suggesting that the LLM indeed guided the agents to discover rewarding states more rapidly, thus mitigating the sparsity challenge.

\bibliographystyle{sbc}
\bibliography{sbc-template}

\end{document}